\documentclass[conference]{IEEEtran}
\IEEEoverridecommandlockouts 

\usepackage{cite}
\usepackage{amsmath,amssymb,amsfonts}
\usepackage{bm}
\usepackage{mathtools}
\usepackage{graphicx}
\usepackage{textcomp}
\usepackage{url}
\usepackage{multirow}
\usepackage{booktabs}
\usepackage{tabularx}
\usepackage{array}
\usepackage{siunitx}
\usepackage[caption=false,font=footnotesize]{subfig}
\newcommand{\ind}[1]{\mathbb{I}\!\left[#1\right]}

\newcommand{\topic}[1]{\path{#1}}

\begin{document}

\title{Robust Global--Local Behavior Arbitration via Continuous Command Fusion Under LiDAR Errors}

\author{
Mohamed Elgouhary and Amr S. El-Wakeel
\thanks{The Authors are with the Lane Department of Computer Science and Electrical Engineering, West Virginia University, Morgantown, WV, USA.
        {\tt\small mae00018@mix.wvu.edu, amr.elwakeel@mail.wvu.edu}}%
\thanks{This work was partially supported by DARPA AI-CRAFT under Grant AWD16069. Also, this work was partially supported by New Frontiers in Research Fund (NFRF) NFRFE-2024-00994. }
}


\maketitle

\begin{abstract}
Modular autonomous driving systems must coordinate global progress objectives with local safety-driven reactions under imperfect sensing and strict real-time constraints. This paper presents a ROS2-native arbitration module that continuously fuses the outputs of two unchanged and interpretable controllers: a global reference-tracking controller based on Pure Pursuit and a reactive LiDAR-based Gap Follow controller. At each control step, both controllers propose Ackermann commands, and a PPO-trained policy predicts a continuous gate $\alpha\in[0,1]$ from a compact feature observation to produce a single fused drive command, augmented with practical safety checks. For comparison under identical ROS topic inputs and control rate, we implement a lightweight sampling-based predictive baseline. Robustness is evaluated using a ROS2 impairment protocol that injects LiDAR noise, delay, and dropout, and additionally sweeps forward-cone false short-range outliers. In a repeatable close-proximity passing scenario, we report safe success and failure rates together with per-step end-to-end controller runtime as sensing stress increases. The study is intended as a command-level robustness evaluation in a modular ROS2 setting, not as a replacement for planning-level interaction reasoning.
\end{abstract}

\begin{IEEEkeywords}
Autoware, ROS2, behavior arbitration, global--local control, LiDAR robustness, sensing impairments, safety validation, reinforcement learning, PPO, modular autonomous driving.
\end{IEEEkeywords}

\section{Introduction}
\IEEEPARstart{B}{ehavior} arbitration is a central requirement in modular autonomous driving stacks. A practical system must balance a global progress objective, such as tracking a planned reference, with local safety-driven reactions to nearby obstacles under imperfect sensing and strict real-time constraints. In practice, perception streams can be corrupted by measurement noise, latency, missing data, and false readings, all of which can degrade decision quality and compromise safety \cite{c_sensor_noise,c_latency,c_fault_injection}. These challenges become more pronounced in interactive driving situations such as close-proximity overtaking, where the vehicle must initiate lateral motion, preserve safe clearance, and return to the desired reference within short decision windows.

On many Ackermann platforms, a high-rate 2D LiDAR scan is a primary input for short-horizon collision avoidance and local maneuvering \cite{c_racing_survey}. However, LiDAR measurements may be affected by range noise, delayed delivery, dropped scans, and false short-range readings \cite{c_sensor_noise,c_latency,c_fault_injection}. When behavior arbitration depends on threshold rules, for example using clearance or relative-distance triggers, such artifacts can induce premature or delayed maneuver initiation, rapid oscillations between behaviors, or overly conservative decisions that reduce progress \cite{c_decision_survey_2021,c_rule_based_sm_2024}.

A practical way to satisfy real-time constraints is to compose simple and well-understood motion behaviors \cite{c_racing_survey}. A global reference-tracking controller, such as Pure Pursuit on a precomputed raceline, is efficient when the path ahead is clear, whereas a reactive LiDAR-based controller, such as Gap Follow, can generate prompt lateral avoidance when the reference is blocked \cite{c_pp,c_stanley,c_raceline,c_followgap,c_racing_survey}. The central difficulty is therefore not the availability of these controllers, but how to coordinate them robustly when the sensing stream is imperfect \cite{c_decision_survey_2021,c_rule_based_sm_2024}.

This paper presents a lightweight arbitration module that learns only the fusion decision while keeping both low-level controllers unchanged and interpretable \cite{c_residual_rl,c_learned_gating}. At each control update, the two controllers propose candidate Ackermann commands, and a reinforcement-learned policy outputs a continuous gate $\alpha\in[0,1]$ that blends them into a single executable action \cite{c_rl_racing,c_hier_rl,c_moe}. Because learning is restricted to a single scalar decision, the resulting interface supports simple safety mechanisms, including an input-staleness watchdog, a minimum-clearance stop, and interaction-mode gating, without modifying the underlying controllers \cite{c_decision_survey_2021}. Although interaction handling can also be addressed at the planning level, our focus here is narrower: coordinating two already-available low-level driving behaviors under impaired LiDAR input while preserving fixed-rate execution and simple runtime overrides. We therefore position the proposed method as a command-level arbitration layer rather than as a replacement for planning-level interaction reasoning.

Beyond nominal task performance, we focus on robustness under controlled LiDAR impairments \cite{c_fault_injection,c_sensor_noise}. We introduce a ROS2-based evaluation setup that applies a fixed base impairment to the LiDAR stream, consisting of Gaussian range noise, scan delay, and scan dropout, and then sweeps the probability of false short-range outliers in the forward cone \cite{c_fault_injection,c_latency}. This produces robustness curves that quantify safe success and failure modes as sensing stress increases. In addition, we report per-step end-to-end controller runtime to characterize real-time stability under the same impaired sensing conditions.

To provide a matched comparison against a model-based alternative, we also implement a lightweight sampling-based predictive baseline that uses the same ROS topic inputs and the same control frequency as the proposed arbitration stack \cite{c_mpc_racing,c_overtake_framework_2023,c_sampling_mpc}. The baseline evaluates a small set of constant-steering kinematic candidates using a reference-tracking objective together with LiDAR-based collision screening \cite{c_sampling_mpc}. Both methods are evaluated in the F1TENTH Gym ROS head-to-head overtaking setting under controlled perception degradations. The resulting study is therefore a simulation-based robustness evaluation of global--local behavior arbitration under controlled LiDAR impairments, with emphasis on both safety outcomes and real-time execution behavior.

\subsection{Contributions}
\begin{enumerate}
  \item \textbf{ROS2-based LiDAR robustness evaluation with runtime characterization:}
  We introduce an impairment-based evaluation protocol that injects LiDAR noise, delay, and dropout and sweeps forward-cone false short-range outliers, yielding robustness curves over task outcomes, unsafe proximity, and per-step end-to-end controller runtime.

  \item \textbf{Matched sampling-based predictive baseline:}
  We implement a lightweight sampling-based predictive baseline that uses constant-steering kinematic candidates with LiDAR-based collision screening and runs at the same control frequency with the same ROS topic inputs as the proposed arbitration stack.

  \item \textbf{Command-level continuous fusion of unchanged controllers:}
  We present a ROS2 arbitration module that fuses Pure Pursuit and Gap Follow through a single continuous gate $\alpha\in[0,1]$ into one Ackermann command at 30\,Hz, augmented with interaction-mode gating, an input-staleness watchdog, and a minimum-clearance override.
\end{enumerate}

\section{Related Work}

\subsection{Geometric tracking for Ackermann platforms}
Lightweight geometric tracking methods remain widely used on Ackermann vehicles because they offer predictable tracking performance at high update rates with low computational cost. Controllers such as Pure Pursuit \cite{c_pp} and Stanley \cite{c_stanley}, often combined with speed profiles along a precomputed reference trajectory, provide strong tracking performance and are straightforward to deploy \cite{c_racing_survey}. In dynamic or interactive settings, however, a purely global tracker may continue following the reference even when the safe driving region changes, for example when another vehicle blocks the path ahead. This limitation motivates arbitration with local safety-oriented behaviors.

\subsection{Reactive LiDAR-based control and collision avoidance}
Reactive collision avoidance methods based on 2D LiDAR are widely used for short-horizon obstacle avoidance because they require limited modeling assumptions and can be executed efficiently. Gap-following and follow-the-gap methods typically enlarge nearby obstacles for safety, identify a feasible free-space direction, and steer toward it \cite{c_followgap,c_racing_survey}. Such controllers are well suited to rapid local maneuver initiation and collision avoidance, but they do not explicitly encode long-horizon progress objectives and may require additional guidance to rejoin a desired reference efficiently after the interaction.

\subsection{Hybrid architectures and arbitration between behaviors}
A common practical architecture combines a global tracker with a local avoidance behavior through switching or blending. Many approaches use mode logic based on thresholds in clearance, time-to-collision, heading error, or relative distance \cite{c_decision_survey_2021,c_rule_based_sm_2024}. While effective in structured settings, these rule-based mechanisms can be sensitive to sensing artifacts: noisy, delayed, or missing measurements may induce frequent switching, excessive conservatism, or delayed maneuver initiation. Alternatives include continuous blending and mixture-of-experts style combinations, in which a scalar or vector gate determines the contribution of each controller at each control step \cite{c_moe,c_racing_survey}. Our approach follows this line, but uses a minimal fusion interface: a single continuous gate over two fixed and interpretable controllers. This design supports straightforward runtime safeguards, such as mode gating and staleness monitoring, while preserving real-time simplicity. It is intended as a command-level coordinator that complements, rather than replaces, planning-level methods.

\subsection{Predictive control and sampling-based approximations}
Model predictive control has been used to balance progress and safety by optimizing short-horizon control sequences subject to kinematic or dynamic constraints \cite{c_mpc_racing,c_overtake_framework_2023}. Predictive methods provide explicit objectives and constraints, but online optimization and collision checking can be computationally demanding, and their performance depends on the accuracy and timeliness of the perception used for candidate evaluation. To reduce computational cost, several works use sampling-based approximations that evaluate a finite candidate set instead of solving a full nonlinear program \cite{c_sampling_mpc}. We use this lightweight form as a matched baseline under identical LiDAR inputs and control rate.

\subsection{Learning-based decision layers and learning-augmented control}
Reinforcement learning has been explored both for end-to-end driving and for higher-level decision layers \cite{c_rl_racing,c_hier_rl}. End-to-end policies can capture complex interaction patterns, but they often require careful safety handling and may be difficult to interpret. Recent work instead uses learning to augment classical controllers through residual RL, learned parameter tuning, or controller selection \cite{c_residual_rl,c_learned_gating}. In this paper, we keep the global tracker and reactive controller unchanged and learn only a single arbitration output that continuously fuses their candidate commands. The main contribution is therefore not the blending idea by itself, but its ROS2 command-level realization together with an impairment-based robustness evaluation.

\subsection{Robustness to sensing imperfections and error-injection evaluation}
Robust autonomous behavior requires reliable operation under sensing imperfections such as noise, latency, missing data, and spurious measurements. Prior work has examined perception uncertainty and robustness through sensor models, delay-aware methods, controlled impairments, and systematic evaluation protocols \cite{c_sensor_noise,c_latency,c_fault_injection}. For LiDAR in particular, false near-range readings and missing scans can strongly affect collision checking and threshold-based mode selection. Our evaluation uses a controlled impairment setup in which base scan degradations are held fixed while forward-cone false short-range outliers are swept to produce robustness curves over both task outcomes and real-time behavior. Sensing uncertainty can be addressed at multiple levels of the autonomy stack; in this work, we specifically evaluate a command-level arbitration interface under controlled LiDAR impairments.

\smallskip

\section{System Overview}
\label{sec:arb}
\begin{figure*}[t]
  \centering
  \includegraphics[width=0.75\textwidth,keepaspectratio]{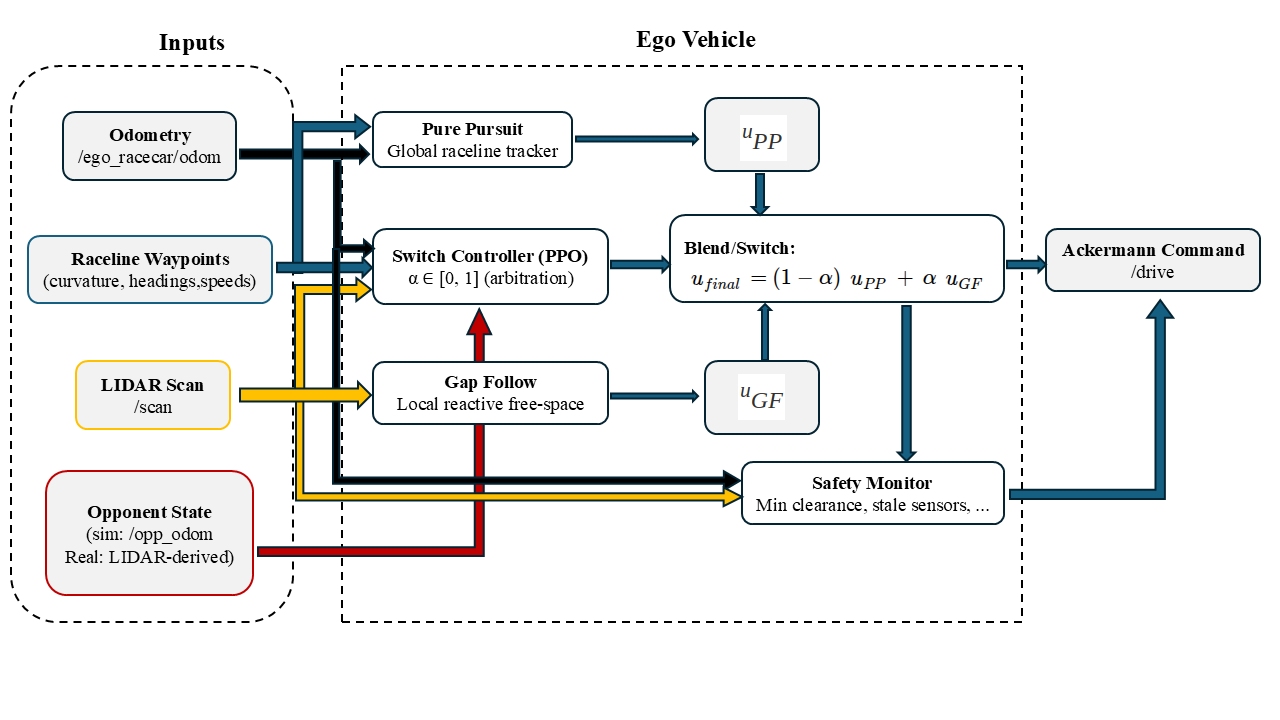}
  \caption{System overview. Pure Pursuit (PP) provides global reference tracking, while Gap Follow (GF) provides local LiDAR-based reactive collision avoidance. A PPO-trained arbiter outputs a continuous gate $\alpha\in[0,1]$ that fuses their candidate Ackermann commands into a single drive command. A safety monitor can override the fused command when clearance or data-health checks fail.}
  \label{fig:sys_over}
  \vspace{-2mm}
\end{figure*}

Fig.~\ref{fig:sys_over} summarizes the proposed global--local behavior arbitration architecture for modular ROS2 driving stacks. In a modular autonomy stack, the proposed module acts as a short-horizon command-level coordinator among existing motion behaviors rather than replacing route planning or trajectory generation. Two low-level controllers run in parallel: a global reference-tracking controller based on Pure Pursuit (PP) and a local reactive collision-avoidance controller based on Gap Follow (GF). At each control step, both controllers propose candidate Ackermann commands, and a learned arbiter outputs a blend weight $\alpha\in[0,1]$ that fuses them into a single drive message. A lightweight safety monitor can override the fused command when minimum-clearance or input-staleness checks are violated.

\subsection{ROS2 Implementation and Topic Interfaces}
\label{sec:ros2_stack}
Table~\ref{tab:ros2_stack} summarizes the ROS2 nodes and topic interfaces used in our experiments. The PP and GF controllers publish candidate drive commands, while the \emph{Behavior Arbiter} node computes features, performs PPO policy inference to obtain the gate, fuses the commands, and publishes the final \topic{/drive} message.

In simulation, the opponent pose is available through odometry and can be used as an optional input during training. To better reflect deployment settings in which opponent pose may not be directly observable, we also evaluate a LiDAR-only variant in which all opponent-pose features are disabled (equivalently, masked with probability $p_{\text{mask}}=1.0$). In that case, the arbiter relies only on LiDAR-derived clearance statistics together with ego and reference-geometry features.

\begin{table}[!t]
\caption{ROS2 stack and topic interfaces used in simulation experiments.}
\label{tab:ros2_stack}
\centering
\footnotesize
\setlength{\tabcolsep}{3pt}
\renewcommand{\arraystretch}{1.15}
\begin{tabularx}{\columnwidth}{@{}l >{\raggedright\arraybackslash}X >{\raggedright\arraybackslash}X@{}}
\toprule
\textbf{Node} & \textbf{Subscribes} & \textbf{Publishes} \\
\midrule
Pure Pursuit (ego) &
\texttt{/ego\_racecar/odom} &
\texttt{/pure\_pursuit\_cmd} \\
\midrule
Gap Follow (ego) &
\texttt{/scan} &
\texttt{/gap\_follow\_cmd} \\
\midrule
Behavior Arbiter &
\texttt{/scan} (or \texttt{/scan\_imp})\newline
\texttt{/ego\_racecar/odom}\newline
\texttt{/opp\_racecar/odom}\newline
\texttt{/pure\_pursuit\_cmd}\newline
\texttt{/gap\_follow\_cmd} &
\texttt{/drive} \\
\midrule
Sampling MPC &
\texttt{/scan}\newline
\texttt{/ego\_racecar/odom} &
\texttt{/drive}\newline
\texttt{/mpc\_ref\_marker}\newline
\texttt{/mpc\_pred\_marker}\newline
\texttt{/mpc\_path\_marker} \\
\midrule
Sensor Impairment &
\texttt{/scan}\newline
\texttt{/opp\_racecar/odom} &
\texttt{/scan\_imp}\newline
\texttt{/opp\_racecar/}\newline
\texttt{odom\_imp} \\
\midrule
RaceEval (metrics) &
\texttt{/scan}\newline
\texttt{/ego\_racecar/odom}\newline
\texttt{/opp\_racecar/odom}\newline
\texttt{/drive} &
JSON logs; summary metrics \\
\bottomrule
\end{tabularx}
\vspace{-2mm}
\end{table}

In our experiments, the arbiter stack and the sampling-based predictive baseline are run in separate heats, since both publish to \topic{/drive}. Therefore, only one high-level controller is active at a time.

\subsection{Low-Level Controllers}
\subsubsection{Pure Pursuit (global reference tracking)}
Pure Pursuit steers toward a lookahead point on a precomputed reference raceline. Let the wheelbase be $L$, the lookahead distance be $L_d>0$, and the lookahead point expressed in the vehicle frame be $(x_d,y_d)$. The commanded curvature and steering angle are
\begin{equation}
\label{eq:pp}
\begin{aligned}
\kappa &= \frac{2y_d}{L_d^{2}} = \frac{2\sin\theta}{L_d},\\
\delta &= \arctan(L\kappa) = \arctan\!\left(\frac{2L\sin\theta}{L_d}\right).
\end{aligned}
\end{equation}
where
\begin{equation*}
\begin{aligned}
\theta &= \operatorname{atan2}(y_d,x_d),\\
L_d &= \sqrt{x_d^2 + y_d^2}.
\end{aligned}
\end{equation*}

\subsubsection{Gap Follow (local reactive collision avoidance control)}
The reactive controller operates directly on the 2D LiDAR scan. It applies a safety bubble around nearby obstacles, identifies a feasible free-space gap in the scan, and steers toward that direction. Forward speed is reduced when free space is limited and increased when the region ahead is clear, enabling rapid local avoidance and maneuver initiation. Because its objective is short-horizon safety rather than long-horizon progress, the vehicle may deviate from the global reference trajectory and may require additional guidance to rejoin it efficiently.

\subsection{Learning-Based Arbitration Module}
\label{sec:rl_arb}

\subsubsection{Problem formulation}
We formulate arbitration as a Markov decision process (MDP) that uses compact LiDAR-derived features rather than raw scans. At each control update $t$ (rate $f_c$), the system constructs an observation vector $s_t\in\mathbb{R}^{d}$ and evaluates two fixed controllers in parallel: a global reference tracker (PP) and a reactive collision-avoidance controller (GF). These controllers produce Ackermann commands
$u_{\mathrm{PP},t}=[\delta_{\mathrm{PP},t},\, v_{\mathrm{PP},t}]^\top$
and
$u_{\mathrm{GF},t}=[\delta_{\mathrm{GF},t},\, v_{\mathrm{GF},t}]^\top$.
The arbiter selects a continuous gate action $a_t=\alpha_t\in[0,1]$ and executes the fused command
\begin{equation}
u_t = (1-\alpha_t)\,u_{\mathrm{PP},t} + \alpha_t\,u_{\mathrm{GF},t},
\label{eq:fusion_vec}
\end{equation}
followed by actuator saturation
$\delta_t\leftarrow \mathrm{sat}(\delta_t;\delta_{\min},\delta_{\max})$
and
$v_t\leftarrow \mathrm{sat}(v_t;v_{\min},v_{\max})$.
This formulation constrains learning to a single scalar decision while preserving the interpretability and nominal stability properties of both low-level controllers.

\subsubsection{Observation encoding}
The arbiter receives a compact feature state $s_t$ built from three sources: (i) forward-cone LiDAR statistics, (ii) reference-geometry summaries such as raceline curvature, and (iii) relative opponent information when available. We use
\begin{equation}
\label{eq:state_vec}
\begin{aligned}
s_t = [&v_t,\;\kappa_t^{(0)},\kappa_t^{(1)},\kappa_t^{(2)},\Delta\kappa_t,\\
&d_{\mathrm{front},t},\; d_{\mathrm{opp},t},\;
\sin\beta_t,\;\cos\beta_t,\;\Delta v_t]^\top .
\end{aligned}
\end{equation}
Here, $\{\kappa_t^{(i)}\}$ summarize upcoming reference curvature and $\Delta\kappa_t$ captures curvature variation over a short lookahead window. The forward-clearance statistic $d_{\mathrm{front},t}$ is computed from beams within a fixed forward field of view using a robust aggregate, such as a minimum or low percentile, to reduce sensitivity to isolated outliers. The opponent terms $(d_{\mathrm{opp},t},\beta_t,\Delta v_t)$ encode relative distance, relative angle, and relative speed.

To handle missing opponent estimates during training, we randomly drop the opponent features. Let $m_t\in\{0,1\}$ be a Bernoulli indicator with $\mathbb{P}(m_t=0)=p_{\mathrm{mask}}$. When $m_t=0$, the opponent features are replaced with fixed default values:
\begin{equation}
\tilde{s}_t = \big[s_t^{\text{(ego)}},\; m_t\,s_t^{\text{(opp)}} + (1-m_t)\,\bar{s}^{\text{(opp)}}\big],
\label{eq:masking}
\end{equation}
where $\bar{s}^{\text{(opp)}}$ is a constant fill value representing a far or unknown opponent state. We additionally apply running mean/variance normalization (VecNormalize) during training to stabilize learning.

\subsubsection{Policy parameterization}
The policy $\pi_\theta(\alpha\mid \tilde{s})$ uses a small multilayer perceptron that outputs a scalar gate constrained to $[0,1]$. In implementation, an unconstrained network output $z_t\in\mathbb{R}$ is mapped to the feasible interval by a sigmoid:
\begin{equation}
\alpha_t = \sigma(z_t)=\frac{1}{1+\exp(-z_t)}.
\label{eq:squash}
\end{equation}
This ensures action feasibility without requiring additional clipping of the gate itself.

\subsubsection{PPO training}
We train the gating policy using Proximal Policy Optimization (PPO) with Generalized Advantage Estimation (GAE) \cite{c_ppo,c_gae}. The policy is a lightweight multilayer perceptron that outputs an unconstrained scalar, which is mapped to the gate $\alpha\in[0,1]$ through Eq.~\eqref{eq:squash}. We use VecNormalize during training and select the final checkpoint from periodic evaluation. Training hyperparameters are summarized in Table~\ref{tab:ppo_hparams}.

\begin{table}[t]
\caption{PPO training hyperparameters.}
\label{tab:ppo_hparams}
\centering
\footnotesize
\setlength{\tabcolsep}{4pt}
\renewcommand{\arraystretch}{1.08}
\begin{tabular}{@{}l l@{}}
\toprule
\textbf{Parameter} & \textbf{Value} \\
\midrule
Total steps & $1.2\times 10^{6}$ \\
Rollout length $n_{\mathrm{steps}}$ & $4096$ \\
Batch size $B$ & $256$ \\
Epochs per update $n_{\mathrm{epochs}}$ & $5$ \\
Discount $\gamma$ & $0.99$ \\
GAE $\lambda$ & $0.98$ \\
Clip range $\epsilon$ & $0.2$ \\
Entropy coef. $c_{\mathrm{ent}}$ & $0.02$ \\
Value-loss coef. $c_v$ & $0.6$ \\
Max grad norm & $0.7$ \\
Target KL & $0.015$ \\
Learning rate & $\eta(\mathrm{frac})=2.4\times10^{-4}\cdot \mathrm{frac}$ (linear) \\
Eval frequency & every $5000$ steps (8 episodes) \\
Checkpoint save & every $25000$ steps \\
\bottomrule
\end{tabular}
\vspace{-2mm}
\end{table}

\subsubsection{Reward design with conservative shaping}
The reward promotes forward progress while penalizing unsafe spacing and rapid changes in the gate. We use
\begin{equation}
\label{eq:reward_generic}
\begin{aligned}
r_t &=
w_{\mathrm{prog}}\,\Delta\mathrm{wp}_t
+ w_v\,v_t
- w_{\mathrm{sm}}\,|\alpha_t-\alpha_{t-1}|\\
&\quad
- w_{\mathrm{risk}}\,\Psi(c_t)
+ r_t^{\mathrm{pass}}
+ r_t^{\mathrm{term}} .
\end{aligned}
\end{equation}
Here, $\Delta \mathrm{wp}_t$ measures reference progress along the raceline, $c_t$ denotes the minimum forward clearance, and $\Psi(\cdot)$ is a barrier-like penalty that increases as clearance approaches a minimum safe radius. The term $r_t^{\mathrm{pass}}$ rewards a completed overtake, while $r_t^{\mathrm{term}}$ applies terminal penalties for collision or leaving the drivable track. Compared with purely reactive objectives, rewarding progress reduces the tendency to remain stalled behind a slower opponent.

To stabilize early learning, we optionally include a conservative reference-gate term. Let $\alpha_t^\star\in[0,1]$ denote a reference gate computed from simple measurements, such as forward clearance and relative direction. We then add an imitation-style penalty with time-varying weight $\lambda_t$:
\begin{equation}
r_t \leftarrow r_t - \lambda_t\,|\alpha_t-\alpha_t^\star|.
\label{eq:reward_shaping}
\end{equation}
Unlike hard-coded switching, this term does not determine the final behavior; it mainly guides exploration toward safer arbitration during early training.

\subsubsection{Runtime smoothing, mode gating, and safety override}
During deployment, we low-pass filter the gate to reduce high-frequency switching:
\begin{equation}
\bar{\alpha}_t = (1-\beta)\bar{\alpha}_{t-1} + \beta\,\alpha_t,\qquad \beta\in(0,1].
\label{eq:alpha_lpf}
\end{equation}
We also restrict the policy to operate only during interaction events. Let $\mathbb{I}_t\in\{0,1\}$ denote an interaction-mode indicator that activates when the forward region is persistently constrained and a leading agent lies within a headway threshold, and deactivates only after sustained clearance for multiple steps. In our evaluation, this interaction corresponds to head-to-head overtaking, although the same gating mechanism can be used in other close-proximity maneuvering scenarios. The executed gate is
\begin{equation}
\alpha^{\mathrm{exec}}_t = \mathbb{I}_t\,\bar{\alpha}_t.
\label{eq:mode_gate}
\end{equation}
Finally, a safety monitor overrides the fused command with a stop if either the forward clearance violates a minimum threshold or required inputs are stale beyond a timeout:
\begin{equation}
u_t \leftarrow [0,\;0]^\top
\quad \text{if} \quad
c_t < c_{\min} \;\; \text{or} \;\; \Delta t_{\mathrm{stale}}>\tau_{\max}.
\label{eq:safety_override}
\end{equation}

\subsubsection{Training protocol in simulation}
Training is performed in the F1TENTH Gym ROS simulator with an ego vehicle and a slower reference-tracking opponent to induce repeated blocked and unblocked interactions. Episodes terminate on collision, leaving the drivable region, or timeout. We train for $1.2\times 10^6$ environment steps with periodic evaluation and checkpointing. PPO hyperparameters are listed in Table~\ref{tab:ppo_hparams}. Training is conducted on the \texttt{Levine} racetrack, and results are reported on the unseen \texttt{IMS} racetrack to evaluate transfer across tracks.

\section{Sampling-Based Predictive Baseline (Sampling MPC)}
\label{sec:mpc}
To compare against a model-based alternative under identical ROS topic inputs and the same control frequency $f_c=30$\,Hz, we implement a lightweight sampling-based predictive baseline. This baseline is intended as a matched local benchmark under the same ROS inputs and control rate as the proposed arbitration stack, rather than as a representative of the full planning literature.

At each control step, the baseline enumerates a small discrete set of constant-steering candidates $\delta^{(m)}$ around a context-dependent center. During standard tracking, this center is aligned with the reference heading; during interaction, the candidate set is expanded toward more aggressive avoidance steering. For each candidate, the controller simulates a short horizon with a kinematic bicycle model to obtain a predicted trajectory $\{p_k\}_{k=0}^{N}$. Candidates are then screened using a conservative LiDAR-based clearance check: if a predicted path enters a region whose measured clearance falls below a safety radius $r_{\text{safe}}$, the candidate is rejected or assigned a large cost. Among feasible candidates, the controller selects the minimum-cost trajectory and executes only its first-step command, repeating the procedure at the next control step in receding-horizon fashion.

We score candidates using a reference-tracking objective with progress and obstacle terms:
\begin{equation}
J = w_{\text{trk}}\sum_{k=0}^{N}\|p_k-p_k^{\text{ref}}\|^2
+ w_\delta\,\delta^2
- w_{\text{prog}}\,\text{prog}
+ w_{\text{obs}}\Phi(c_{\min}),
\end{equation}
where $p_k$ is the candidate position, $p_k^{\text{ref}}$ is the raceline reference, $\text{prog}$ denotes forward progress over the horizon, and $\Phi(c_{\min})$ penalizes small minimum clearance $c_{\min}$ along the predicted path; colliding candidates receive a large penalty.

Table~\ref{tab:sampling_mpc_params} lists the sampling-based predictive baseline settings. The cost weights are mode-dependent, prioritizing obstacle clearance during interaction and reference tracking during standard motion.

\begin{table}[!t]
\caption{Sampling-based predictive baseline parameters.}
\label{tab:sampling_mpc_params}
\centering
\footnotesize
\setlength{\tabcolsep}{4pt}
\renewcommand{\arraystretch}{1.05}
\begin{tabular}{@{}l l@{}}
\toprule
\textbf{Setting} & \textbf{Value} \\
\midrule
Horizon steps $N$ & $8$ \\
Timestep $\Delta t$ & $0.1$\,s \\
Candidates $M$ (standard tracking) & $9$ \\
Candidates $M$ (interaction) & $17$ \\
Steering limits $\delta$ & $[-0.4189,\,0.4189]$ rad \\
Safety radius $r_{\text{safe}}$ & $0.55$\,m \\
\bottomrule
\end{tabular}
\vspace{-2mm}
\end{table}

\section{Robustness Evaluation Protocol}
\label{sec:eval}

\subsection{Simulation Setup}
All experiments are conducted in the F1TENTH Gym ROS simulator to evaluate the controllers under controlled LiDAR impairments. Training is performed on the \texttt{Levine} track, while robustness evaluation is reported on the unseen \texttt{IMS} track to assess transfer across environments. For each configuration, we run $N_{\text{heats}}=10$ interaction heats per random seed. The ego vehicle is controlled either by the proposed global--local arbitration stack or by the sampling-based predictive baseline, with only one high-level controller active per heat. The opponent is a lower-speed reference-tracking agent, implemented as Pure Pursuit with a reduced target speed relative to the reference speed profile, to induce repeated blocked and unblocked interactions. Both methods consume the same ROS topic inputs and execute at the same control frequency $f_c=30$~Hz.

\subsection{Metrics}
A heat is considered \emph{successful} if the ego completes at least one pass within the time limit without collision or leaving the drivable track. We report the following metrics:

\begin{itemize}
  \item \textbf{Success Rate (SR):}
  Let $N_{\text{succ}}$ be the number of successful heats and $N$ the total number of heats. Then
\begin{equation}
  \mathrm{SR}=\frac{N_{\text{succ}}}{N}.
\end{equation}

  \item \textbf{Safety outcomes:}
  collision rate (\textbf{Coll.}), off-track termination rate (\textbf{OffTrk}), and unsafe-proximity rate (\textbf{Unsafe}).
  Let $c_t$ denote the minimum front-sector clearance at control step $t$, and let
  $c_{\min}=\min_t c_t$ over the heat. We set $c_{\text{unsafe}}=0.35$~m and define an unsafe-proximity event using a persistence parameter $M$ (in control steps):
  \begin{equation}
    \text{Unsafe}=\ind{\,\exists\,t:\;\bigwedge_{i=0}^{M-1}\big(c_{t+i}<c_{\text{unsafe}}\big)\,}.
  \end{equation}
  We use $M{=}3$ control steps to reduce sensitivity to isolated sensing outliers.

  \item \textbf{Timeout rate (T/O):}
  the fraction of heats that end by reaching the time limit before a successful pass.

  \item \textbf{Real-time compute:}
  per-step end-to-end controller runtime $T_{\text{rt}}$ (ms), measured from receiving the latest observation to publishing the drive command. We report the mean runtime (\textbf{Mean}) and worst-case runtime (\textbf{Worst}) over each heat.
\end{itemize}

\subsection{LiDAR Impairments}
To evaluate robustness to LiDAR imperfections in a controlled and consistent setting, we apply impairments at evaluation time using a ROS2 LiDAR-impairment node that modifies the scan stream before it reaches the controller. Unless stated otherwise, all runs use a fixed \emph{base scan impairment} consisting of:
(i) additive zero-mean Gaussian range noise with $\sigma=0.05\,\mathrm{m}$, clipped to $[r_{\min}, r_{\max}]$;
(ii) a scan delay of $\tau_{\text{scan}}=200\,\mathrm{ms}$ implemented via a FIFO queue; and
(iii) scan dropout with probability $p_{\text{drop}}=0.3$ per scan using zero-order hold, i.e., republishing the last valid scan.

In addition to the base impairment, we inject front-facing false short-range outliers to emulate occasional false close readings that can corrupt collision checks and free-space estimates. With probability $p_{\text{out}}$ per scan, we select a fraction $f_{\text{out}}=0.12$ of beams within a symmetric $40^\circ$ forward field of view and set their ranges to $r_{\text{out}}=0.10\,\mathrm{m}$, followed by clipping to sensor limits. We sweep
$p_{\text{out}} \in \{0.0,\,0.2,\,0.4\}$ while keeping the other impairment parameters fixed. Each configuration is repeated over $S=3$ random seeds with $N_{\text{heats}}=10$ heats per seed, yielding robustness curves that relate safety, task success, and runtime behavior to increasing sensing stress.

\begin{table}[t]
\caption{Experimental design: LiDAR impairment parameters, repeated trials, and runtime measurement setup.}
\label{tab:exp_design}
\centering
\footnotesize
\setlength{\tabcolsep}{3.0pt}
\renewcommand{\arraystretch}{1.12}
\begin{tabularx}{\columnwidth}{@{}l X@{}}
\toprule
\textbf{Item} & \textbf{Setting} \\
\midrule
Simulator & F1TENTH Gym ROS (ROS2) \\
Tracks & Train: \texttt{Levine}; Eval: unseen \texttt{IMS} \\
Control rate & $f_c=30$~Hz ($T_s=33.3$~ms) \\
\midrule
Base LiDAR range noise & $\sigma=0.05$~m (zero-mean Gaussian; clipped to $[r_{\min}, r_{\max}]$) \\
Base LiDAR scan delay & $\tau_{\text{scan}}=200$~ms (FIFO queue) \\
Base LiDAR dropout & $p_{\text{drop}}=0.3$ per scan (zero-order hold / last-scan republish) \\
\midrule
Outlier injection region & Forward cone FOV $=40^\circ$ (symmetric about forward axis) \\
Outlier beam fraction & $f_{\text{out}}=0.12$ of beams in the forward cone \\
Outlier range value & $r_{\text{out}}=0.10$~m (then clipped to sensor limits) \\
Outlier probability sweep & $p_{\text{out}} \in \{0.0,\;0.2,\;0.4\}$ per scan \\
\midrule
Repetitions & $S=3$ random seeds; $N_{\text{heats}}=10$ heats per seed per configuration \\
Opponent behavior & Pure Pursuit raceline follower at reduced target speed (fixed across methods) \\
\midrule
Runtime measurement & Per-step end-to-end controller runtime, measured from receiving the latest state/observation to publishing the drive command \\
Compute platform & Dell Precision 3660; 13th Gen Intel Core i9-13900 (32 threads), 32\,GiB RAM; Ubuntu 22.04.5 LTS (64-bit); NVIDIA GPU + Intel(R) Graphics (RPL-S) \\
\bottomrule
\end{tabularx}
\vspace{-2mm}
\end{table}

Table~\ref{tab:exp_design} summarizes the impairment parameters, repetition protocol, and compute platform used for runtime measurements.

\section{Results}
\label{sec:results}
\normalcolor

\begin{figure}[t]
  \centering
  \includegraphics[width=0.85\columnwidth]{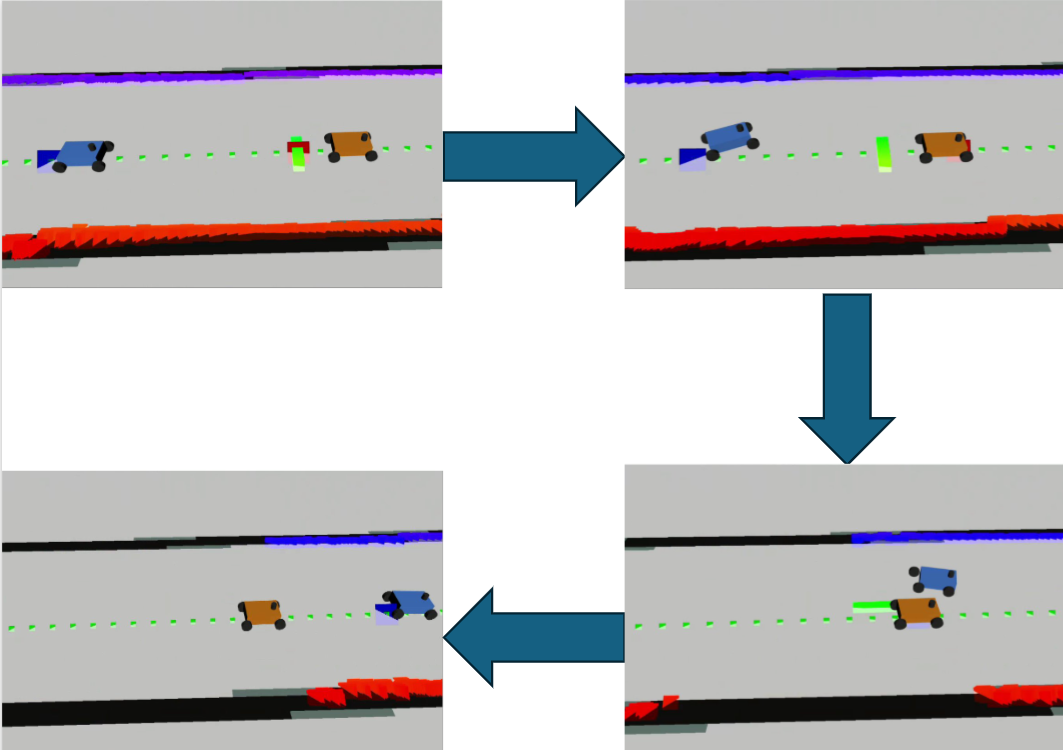}
  \caption{Qualitative interaction sequence in simulation. The ego vehicle transitions from global reference tracking to a local free-space maneuver to pass the leading vehicle, then returns to reference tracking.}
  \label{fig:overtake-seq}
  \vspace{-2mm}
\end{figure}

Fig.~\ref{fig:overtake-seq} shows a representative passing sequence under the proposed arbitration strategy.

\subsection{Baseline Performance and Real-Time Stability}
Under standard sensing on \texttt{IMS}, the proposed global--local arbitration module achieved a $100\%$ success rate over $N_{\text{heats}}=10$ interaction heats, with zero collisions, zero off-track terminations, and zero timeouts (Table~\ref{tab:Baseline}). The sampling-based predictive baseline achieved $90\%$ success, with one heat ending in timeout.

In addition to task outcomes, we report per-step end-to-end controller runtime to characterize real-time stability. While the sampling-based predictive baseline attains a lower mean runtime (24.15~ms versus 31.95~ms), it exhibits larger worst-case runtime spikes (61.83~ms versus 47.12~ms). Both methods execute at a fixed control rate of $f_c=30$~Hz; the reported runtime reflects sensing-to-command processing latency per update, rather than the target control period itself ($\approx 33.3$~ms). In interactive scenarios, rare slow updates can reduce effective responsiveness during short decision windows, such as when initiating a lateral maneuver, which motivates reporting both mean and worst-case runtime.

\begin{table}[t]
\caption{Baseline interaction performance on \texttt{IMS} in F1TENTH Gym ROS ($N_{\text{heats}}=10$).}
\label{tab:Baseline}
\centering
\footnotesize
\setlength{\tabcolsep}{2.2pt}
\renewcommand{\arraystretch}{1.15}
\resizebox{\columnwidth}{!}{%
\begin{tabular}{l c c c c c c}
\toprule
Method & SR (\%) & Coll. (\%) & OffTrk (\%) & T/O (\%) & Mean runtime (ms) & Worst runtime (ms)\\
\midrule
Arbiter (PP+GF) & 100.0 & 0.0 & 0.0 & 0.0 & 31.95 & 47.12 \\
Sampling MPC & 90.0 & 0.0 & 0.0 & 10.0 & 24.15 & 61.83 \\
\bottomrule
\end{tabular}%
}
\end{table}
\normalcolor

\subsection{Robustness Curves Under LiDAR Impairments}
\begin{figure}[t]
  \centering
  \includegraphics[width=0.85\columnwidth]{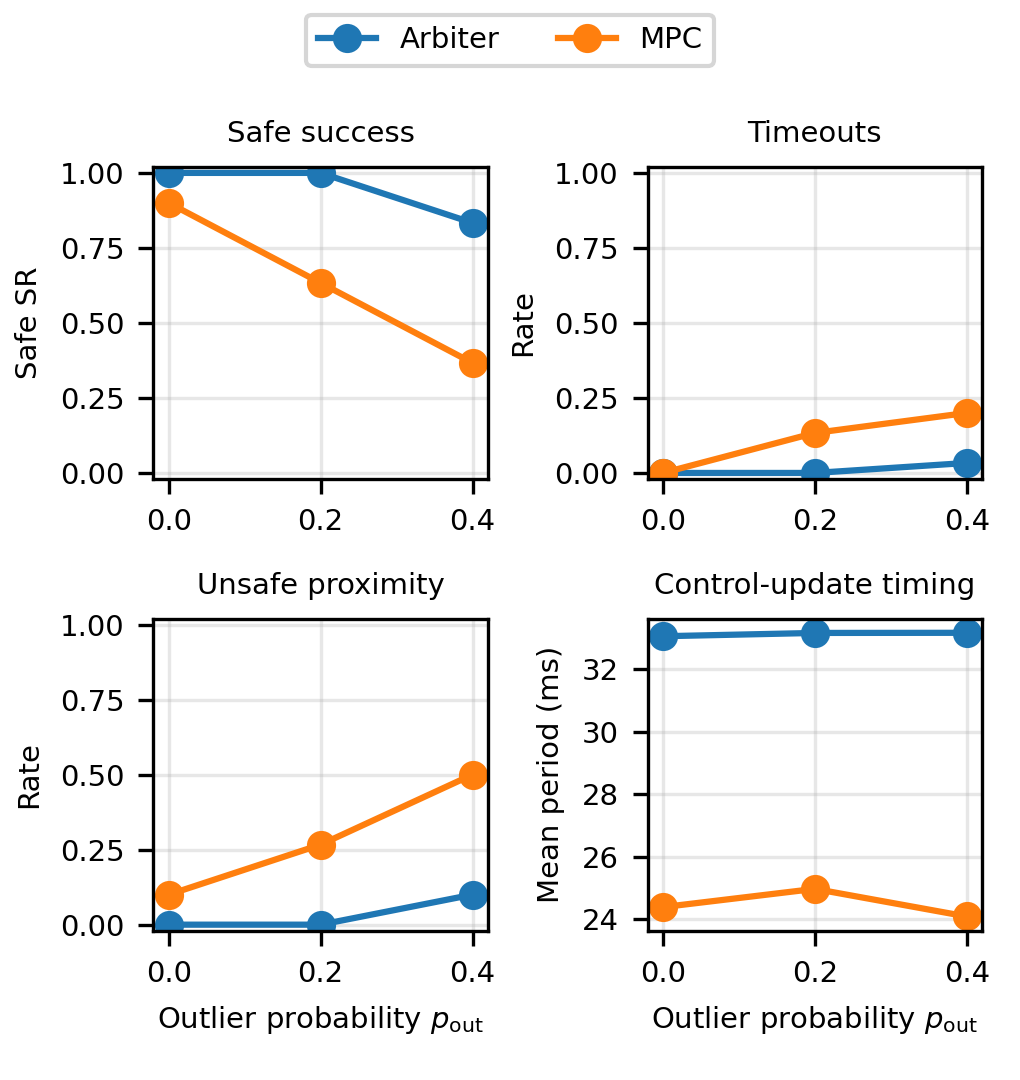}
  \caption{Robustness under base LiDAR impairments (noise, 200\,ms delay, 0.3 dropout) with front-cone outliers swept over $p_{\text{out}}\in\{0,0.2,0.4\}$.}
  \label{fig:robust}
  \vspace{-2mm}
\end{figure}

Fig.~\ref{fig:robust} summarizes robustness to front-cone false short-range outliers under the fixed base scan impairments of noise, delay, and dropout. We report (i) safe success rate (SR$_{\text{safe}}$), (ii) failure-mode rates including timeout and unsafe proximity, and (iii) mean controller runtime, to show how safety, task completion, and real-time behavior evolve as sensing stress increases.

\subsection{Discussion}
Across all outlier levels, the arbitration module exhibits a slower degradation in SR$_{\text{safe}}$ than the sampling-based predictive baseline. The failure breakdown further shows that, as $p_{\text{out}}$ increases, the baseline fails primarily through timeouts or unsafe-proximity events. This behavior is consistent with its reliance on the current LiDAR scan to screen candidate trajectories for collision risk and to evaluate candidate cost.

False short-range outliers make the forward region appear artificially blocked. As a result, the sampling-based predictive baseline may reject more candidate trajectories as unsafe and may select overly cautious, low-progress actions. Scan delay and dropout can amplify this effect: when candidate screening depends on stale or repeated scans, the controller can remain conservative for multiple update cycles, which increases the likelihood of timeout.

In contrast, the arbitration module fuses two stable low-level behaviors and bases its decision on a compact set of summary features, such as forward free-space statistics, rather than checking every LiDAR beam along each predicted trajectory. This aggregation, together with gate smoothing and context-dependent activation, likely reduces sensitivity to isolated outliers and limits rapid oscillations between behaviors, leading to a more gradual decline in safe success under the same sensing impairments.

The runtime results highlight an additional real-time distinction. Although the sampling-based predictive baseline has a lower mean runtime, it exhibits larger worst-case runtime spikes even under standard sensing conditions. Such spikes can reduce effective control responsiveness during short critical windows. By comparison, the arbiter performs a single policy evaluation followed by a linear blend, resulting in steadier per-step end-to-end controller runtime and more consistent closed-loop behavior when measurements are delayed, dropped, or temporarily corrupted.

These results should be interpreted within the scope of the present evaluation. The study considers one repeatable close-proximity overtaking scenario in simulation and therefore does not establish generalization to richer interaction settings or real-vehicle operation. It also does not isolate the effects of gate smoothing, interaction-mode gating, reference-gate shaping, opponent-feature masking, or the safety override. Broader comparisons, larger heat counts, and dedicated component-wise ablations remain important directions for future work.

\section{Conclusion}
This paper evaluated a reinforcement-learned global--local behavior arbitration module based on continuous command fusion and compared it against a sampling-based predictive baseline in ROS2 simulation using F1TENTH Gym ROS. The proposed method learns a single continuous fusion gate while preserving two interpretable low-level behaviors: a global reference-tracking controller and a local LiDAR-based collision-avoidance controller. This design supports reliable interaction handling at $30$~Hz together with consistent per-step end-to-end controller runtime.

Under controlled LiDAR impairments, including Gaussian range noise, scan delay, scan dropout, and a sweep of front-cone false short-range outliers, the arbitration module maintained higher safe success and lower failure rates than the sampling-based predictive baseline. While the baseline remained competitive under standard sensing conditions, its performance degraded more sharply as outlier probability increased, consistent with its reliance on horizon-wise scan-based collision screening and candidate evaluation under impaired or stale measurements.

Overall, these results suggest that continuous global--local command fusion can serve as a practical command-level coordination mechanism for modular autonomy stacks under the sensing impairments studied here. Future work will extend the evaluation to additional interaction settings, broader baselines, component-wise ablations, and real-vehicle or real-sensor validation in Autoware-oriented workflows.

\bibliographystyle{IEEEtran}
\bstctlcite{BSTcontrol}
\bibliography{refs}

\end{document}